%% file: main.tex
\documentclass[11pt]{article}
\input{layout}
\hypersetup{pdftitle={Separating Action Allocation from Image Matching in Visual Token Routing},pdflang={en}}
\begin{document}
\papertitle{Separating Action Allocation from Image Matching\\in Visual Token Routing}
\begin{abstract}
Adaptive visual token routers aim to save computation while preserving the image detail needed to answer each question. A higher average score could reflect better choices for individual images or more frequent use of high resolution on tasks that need it. We distinguish the two by keeping action counts fixed within each task and comparing the router's assignments with random pairings of the same actions and images. On equally weighted ChartQA and TextVQA, both Qwen2.5-VL-3B and -7B gain from matching but lose more from their action frequencies relative to a task table, with a 1.71-point gain and a 3.95-point loss on 7B. Exchanging the continuous resolutions chosen by the external selector CARES also lowers quality on 7B. When requests arrive with task labels, the table outscores the discrete router by 1.92 points and saves 31.76\,ms per complete 7B request on average.
\end{abstract}

\section{Introduction}
A chart can support two rather different questions. Identifying whether a curve rises may take only a coarse view. Reading a decimal on its axis can require the original resolution. Higher resolution gives a vision--language model more visual tokens to process~\cite{qwen,cares}. A service that answers many such questions must decide how often to pay for fine detail and which requests need it.

Image-conditioned policies choose a resolution or a pruning strategy for each request~\cite{cares,vip}. In a service that already distinguishes charts from scene text, a simpler policy can reserve a different budget for each task without inspecting the image. Giving high resolution to more charts changes the average score even if the recipients are drawn at random. An image router can also identify relatively safe charts to compress and still compress too many. The same aggregate comparison can conceal either effect.

\begin{figure}[htbp]\centering
\includegraphics[width=.985\linewidth]{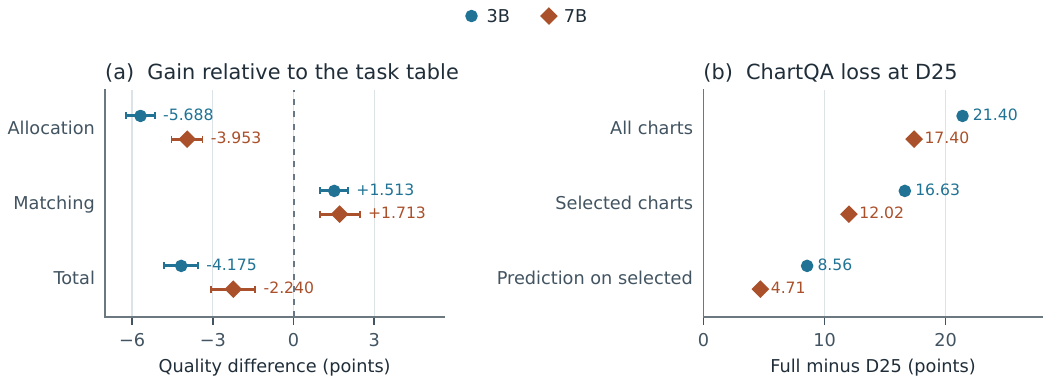}
\caption{\textbf{The router selects safer charts to compress and still compresses too many.} (a) Quality relative to the task table on equally weighted ChartQA and TextVQA. Intervals are pointwise 95\% paired source-bootstrap intervals with the frequencies recomputed in each draw. (b) Full-to-D25 quality loss on ChartQA. The prediction and selected-image measurement use the same charts.}\label{fig:matching}\end{figure}

We make both comparisons on the same images. First redistribute the router's actions at random within each task, preserving the number of uses of every action. The change from random to actual pairing measures matching. The difference between this random control and a task table measures allocation. On charts the two effects have opposite signs. The router selects images that lose less quality under downsampling than the average chart, yet even for these images it predicts too small a loss (Figure~\ref{fig:matching}).

Three experiments connect the decomposition to decisions a service can make. Answers for all seven actions on each confirmation image give both terms on the observed sample and bound how much better the pairing could be. Real inference after exchanging the resolutions recommended by CARES tests matching outside the discrete menu. Running four complete workflows on identical requests then reveals how much downstream time a preview recovers. We also test when a task label alone improves quality at a common target-action time. A retrospective comparison also asks whether choosing hard images with a different model changes which compressor appears to perform better.

\section{What can a routing comparison identify?}
\subsection{What existing comparisons hold fixed}
VTC-Bench identifies images on which downsampling destroys detail needed for an answer~\cite{vtc}. PAA-Bench tests whether selected tokens preserve a model's answer better than random tokens from the same image and measures speed at a common preservation target~\cite{paa}. Song et al. examine whether within-image token selections are stable under meaning-preserving perturbations~\cite{reliability}. VIP-Router compares a learned strategy with fixed strategies and an oracle free to choose its best strategy on every image~\cite{vip}. ET-Prune varies token budgets with the evidence required by text-rich inputs~\cite{etprune}. An oracle free to change action counts can improve both the pairing and the allocation.

Our control exchanges whole actions across images in the same task, preserving their counts. For CARES the exchanged quantity is its recommended long-edge size~\cite{cares}. Quality--cost frontiers and random mixtures have also been used to evaluate LLM routing~\cite{routerbench,misguided}. We use a common mixture across tasks to measure the value of task information, and reassignment within each task to measure the value of image information at fixed action frequencies.

\subsection{A random assignment and an upper bound at the same counts}
Suppose that we have run every candidate action on each image--question pair. Denote the resulting quality by $q_{ia}\in[0,100]$ for source $i$ and action $a$. The policy assigns weight $w_{ia}$ to that action, with $\sum_a w_{ia}=1$. For a deterministic router only the chosen action has weight one. With $n_t$ sources in task $t$ and a fixed task weight $\pi_t$, the policy score and its task-wise action frequencies are
\begin{equation}
 Q(w)=\sum_t\frac{\pi_t}{n_t}\sum_{i\in t}\sum_a w_{ia}q_{ia},
 \qquad p_{ta}=\frac{1}{n_t}\sum_{i\in t} w_{ia}.
 \label{eq:means}
\end{equation}
A task table $v$ may select a single action per task or randomize according to a task-specific mixture. To remove image information from a deterministic router, permute its action labels uniformly within each task. Each source then receives action $a$ with probability $p_{ta}$. Its expected score is therefore obtained by setting $w_{ia}=p_{ta}$ for every $i\in t$ in Eq.~\ref{eq:means}. Call this score $Q(p)$. With the complete action matrix we can compute it directly, without sampling permutations. Averaging over permutations of the weight vectors gives the same expression for fractional assignments. Adding and subtracting $Q(p)$ yields
\begin{equation}
 Q(w)-Q(v)=\underbrace{Q(p)-Q(v)}_{A:\ \text{action allocation}}
           +\underbrace{Q(w)-Q(p)}_{M:\ \text{image matching}}.
 \label{eq:decomp}
\end{equation}
Both terms use the same answers. For a deterministic router and a task table with action $b_t$, define $d_{ia}=q_{ia}-q_{ib_t}$. Action $a$ contributes $p_{ta}\mathbb{E}_t[d_{ia}]$ to allocation and $p_{ta}(\mathbb{E}_t[d_{ia}\mid w_{ia}=1]-\mathbb{E}_t[d_{ia}])$ to matching before task weighting. A model may therefore recognize inputs with a smaller loss and still overuse a lossy action.

For a deterministic router, how much of the available matching gain has it captured? Give each action one slot for each observed use. Maximize $\sum_{t,i,a}(\pi_t/n_t)x_{ia}q_{ia}$ over binary assignments with one action per source, $\sum_a x_{ia}=1$, and precisely the observed slot counts, $\sum_{i\in t}x_{ia}=n_tp_{ta}$. The maximum-weight assignment gives $Q^*_{\rm quota}$. We report the share $M/(Q^*_{\rm quota}-Q(p))$ whenever the denominator is positive. This oracle can read all evaluation answers. It bounds quality at fixed counts. The actual time can change with the recipient image and generated answer.

\subsection{When is a task label enough, and when does a preview pay?}
Let $\mu_{ta}$ be mean quality and $c_{ta}$ mean target-action time for task $t$ under action $a$. Let $C$ denote the task-weighted mean of these costs for a policy. On representative development data we enumerate task tables and choose the highest-scoring feasible table under a budget. A shared baseline uses the same action mixture for every task. At a nonnegative time price $\lambda$, the population optima obey
\begin{equation}
 \sum_t\pi_t\max_a(\mu_{ta}-\lambda c_{ta})
 \ \geq\ \max_a\sum_t\pi_t(\mu_{ta}-\lambda c_{ta}).
 \label{eq:task}
\end{equation}
\emph{Proof.} Take a maximizing action on the right. Its priced value in each task is bounded by that task's maximum. Summing proves the inequality. Equality holds exactly when a common action is optimal in every positive-weight task. A common random mixture cannot increase a linear priced objective. For the corresponding budget guarantee, suppose table $b$ attains the left side at price $\lambda$ and let $B=C(b)$. Then any task policy $v$ with $C(v)\leq B$ satisfies $Q(v)-\lambda C(v)\leq Q(b)-\lambda B$ and hence $Q(v)\leq Q(b)$. These statements use the same population and costs throughout. The held-out performance of a table estimated from finite data remains an empirical question.

These quantities correspond to different clocks. Let $c_{ia}$ be the time to execute the chosen target action and $h_i(w)$ the remaining request overhead, including any preview and dispatch. The mean complete time $T(w)$ averages $\sum_a w_{ia}c_{ia}+h_i(w)$ over the workload. To assess the value of task information we compare task tables and shared mixtures at a common target-action time. To decide whether to run an image preview, we charge the entire request. The development search can enable preview for charts, scene text, both or neither.

Let $\Delta T=T(w)-T(v)$ use complete-request times on the same source distribution as $Q$. At price $\lambda$ quality points per unit of time, Eq.~\ref{eq:decomp} gives the break-even condition
\begin{equation}
 U_\lambda(w)>U_\lambda(v)\quad\Longleftrightarrow\quad M>-A+\lambda\Delta T,
 \qquad U_\lambda=Q-\lambda T .
 \label{eq:breakeven}
\end{equation}
This makes the deployment test explicit. If allocation loses quality, matching must first make up that loss. Any net increase in complete time raises the required matching gain by $\lambda\Delta T$. We measure the quality terms on the confirmation matrix and evaluate complete utility on paired, timed requests using their own answers.

\section{Experimental design}
\paragraph{Workload and action menu.}
The primary workload gives ChartQA and TextVQA equal weight. ChartQA uses relaxed correctness and TextVQA official soft consensus. We report GQA answer accuracy separately~\cite{chart,textvqa,gqa}. All quality scores range from 0 to 100. One source is one image and its question. On Qwen2.5-VL-3B-Instruct and -7B-Instruct~\cite{qwen}, Full is the original unpruned processor setting. D25 and D09 reduce input resolution to target $1/4$ and $1/9$ of the Full visual tokens. DivPrune~\cite{divprune} and VisPruner~\cite{vispruner} each prune at the same two target fractions, giving seven actions. We use identical model weights, prompts and decoding settings across actions. Rounding in the processor affects the realized token counts.

\paragraph{Development, confirmation and reuse of sources.}
For each of three tasks, the 3B development split has 2,400 base-training, 600 adaptation, 400 budget-selection and 600 locked diagnostic sources. The 7B model uses 600 adaptation and 400 budget-selection sources per task. Fresh 3B confirmation data contain 2,000 charts, 2,000 scene-text images and 1,000 GQA images. The 7B confirmation set uses 1,000 per task drawn from those same pools. Thus 5,000 distinct confirmation images give 8,000 model--source pairs. Running seven actions and default CARES on each pair produces 64,000 answers. Table~\ref{tab:scale} gives separate counts for the resolution intervention and repeated timing sessions.

\paragraph{What each policy knows at decision time.}
Both task tables read the task label at entry and use the same development menu and budget. The population table $\su$ is selected on all development sources. The sensitive table $\sa$ is selected on compression-sensitive sources. On TextVQA, sensitivity requires full credit under Full and zero under D25. This outcome-defined label is available only for offline analysis. Both tables select Full for charts. For scene text, $\su$ selects D25, while $\sa$ mixes Full and D09, using Full on 22.15\% (3B) and 26.20\% (7B) of confirmation sources.

Our controlled image router $R$ previews a 224-pixel crop and up to 77 question tokens using an independent CLIP ViT-B/32 encoder~\cite{clip}. It uses a seven-output quality update and a development-fitted time price to choose an action without seeing the target answer. The external policy CARES uses its released default weights and processor ($\beta=0$) to recommend a continuous resolution. We fixed the primary comparisons, policies and seed before confirmation scoring. The subsequent budget-grid study is exploratory.

\section{Matching gains and allocation losses}
\subsection{The discrete router matches images well and overuses downsampling}
Start with the three assignments that use exactly the same counts in Table~\ref{tab:quota}. On 7B, random pairing scores 81.147, the router scores 82.860, and an oracle that fills the same slots scores 90.250. Matching thus adds 1.713 points (95\% paired interval $[0.973,2.460]$), or 18.82\% of the available gain. Changing from the population table's frequencies to the router's costs 3.953 points ($[-4.536,-3.382]$ for the signed change). The net difference is $-2.240$ ($[-3.070,-1.420]$). On 3B, matching adds 1.513 ($[0.998,2.025]$), allocation loses 5.688 ($[-6.213,-5.155]$) and the total is $-4.175$ ($[-4.825,-3.549]$). The 3B router uses 19.94\% of its available matching gain.

\begin{table}[htbp]\centering\small
\caption{\textbf{Quality at the router's action counts.} The first three assignments share the same task-wise counts. The population table has its own allocation. Scores average ChartQA and TextVQA equally over 4,000 primary sources on 3B and 2,000 on 7B. The last row expresses observed matching gain as a share of the gain available above random assignment.}\label{tab:quota}
\begin{tabular*}{.86\linewidth}{@{\extracolsep{\fill}}lrr@{}}\toprule
\textbf{Assignment or statistic} & \textbf{3B} & \textbf{7B}\\\midrule
Random within task, $Q(p)$ & 74.422 & 81.147\\
Observed router, $Q(R)$ & 75.935 & 82.860\\
Optimal pairing at fixed counts, $Q^*_{\rm quota}$ & 82.010 & 90.250\\\midrule
Population task table, $Q(\su)$ & 80.110 & 85.100\\\midrule
Headroom used, $M/[Q^*_{\rm quota}-Q(p)]$ & 19.94\% & 18.82\%\\\bottomrule
\end{tabular*}\end{table}

The chart data show where the signs come from (Figure~\ref{fig:matching}b). On 7B, downsampling all charts to D25 would cost 17.40 points relative to Full. On the 416 of 1,000 charts actually assigned D25 the loss is smaller, 12.02 points. This difference accounts for $+2.24$ matching points within ChartQA. Yet using D25 this frequently contributes $-7.24$ allocation points, and the router expected a loss of only 4.71 on the selected charts. Of all 461 charts receiving any action other than Full, 57 lose quality and 404 tie. The 3B router sends 980 of 2,000 charts to D25. The loss falls from 21.40 on all charts to 16.63 on selected charts, still well above the predicted 8.56. In both models the selected group is safer to compress, and the loss on that group is larger than forecast.

Decision cost also affects action frequencies. We hold the router's seven quality predictions fixed and rerun development selection with the preview priced as cheaply as a lookup. Compared with the original $R$, these new choices improve confirmation quality by 4.175 points on 3B and 2.435 on 7B. Restoring the original preview charge makes their \emph{predicted} complete time respectively 57.50 and 52.54\,ms higher than $R$'s. The counterfactual shows that a cheaper decision leaves room for more target computation. Its effect depends on the same quality predictions that underestimate chart losses, so the two diagnostics do not supply separate causal percentages.

To test whether a simple calibration fixes the problem, we estimate the mean observed-minus-predicted score for each task and action on 800 representative development images per model and add that offset to future predictions. For any pair of actions this preserves the across-image ranking of their predicted difference. After refitting the time price to the original development budget, the confirmation change is $-0.025$ points on 3B ($[-0.202,0.152]$) and $-0.600$ on 7B ($[-1.100,-0.100]$). Correcting the average error alone fails to recover the allocation loss in this experiment.

\subsection{Does a published resolution selector match its images?}
We run the released CARES decision path, then permute its recommended long-edge sizes within each task. Each of five prespecified bijections preserves the task's multiset of sizes. We resize the recipient image and regenerate the answer for each assignment, giving 3,000 new answers per model on 300 charts and 300 scene-text sources. Image aspect ratios, processor rounding and response lengths can change the realized token or time cost after permutation.

\begin{figure}[htbp]\centering
\includegraphics[width=.985\linewidth]{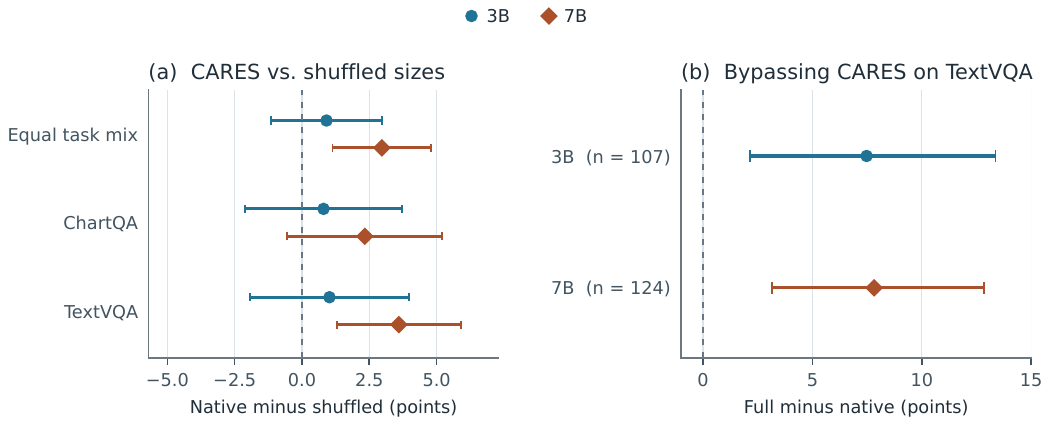}
\caption{\textbf{Two interventions on CARES.} (a) Native minus shuffled resolution on 600 sources per model. The 95\% intervals pair sources conditional on the five within-task permutations. (b) Full minus CARES on timed TextVQA sources assigned Full in every sensitive-table repeat. The 95\% intervals resample eligible sources and sessions. Positive quality favors the first policy in each contrast.}\label{fig:external}\end{figure}
The native assignment improves on shuffled sizes by 2.97 points on 7B ($[1.13,4.80]$) and 0.91 on 3B ($[-1.16,2.98]$). The 7B advantage is clearer on TextVQA ($+3.60$, $[1.31,5.89]$) than on ChartQA ($+2.33$, $[-0.54,5.21]$). Holding sources fixed and estimating Monte Carlo uncertainty across the five draws gives aggregate intervals of $[-0.89,2.71]$ on 3B and $[1.37,4.56]$ on 7B. Thus the data resolve a matching gain for 7B, while allowing either direction on 3B.

A second intervention asks whether the service should invoke the selector for these requests at all (Figure~\ref{fig:external}b). The frozen sensitive table assigned Full in every repeat to 107 of 500 timed text images on 3B and 124 on 7B. On these exact sources Full outscores CARES by 7.48 ($[2.15,13.36]$) and 7.82 points ($[3.15,12.82]$), and saves 626.44 ($[602.77,654.01]$) and 574.04\,ms ($[546.39,601.80]$), respectively. Full also scores higher on the entire TextVQA confirmation pool (85.67 vs.\ 79.42 on 3B and 92.79 vs.\ 85.89 on 7B). The paired latency estimates apply to the timed subsets.

\section{Workload composition can reverse the preferred allocation}
The two task tables ask a different allocation question: which population should select the action frequencies? Both use Full on charts, so all of their quality difference comes from scene text. On TextVQA the table selected on all development sources beats the table selected on sensitive sources by 7.25 points on 3B and 6.71 on 7B. At equal task weights this yields $\su-\sa=+3.625$ ($[2.670,4.580]$) and $+3.355$ ($[2.128,4.582]$). To see which service workloads would favor the other table, we change only the prevalence of sensitive text images.

\begin{figure}[htbp]\centering
\includegraphics[width=.985\linewidth]{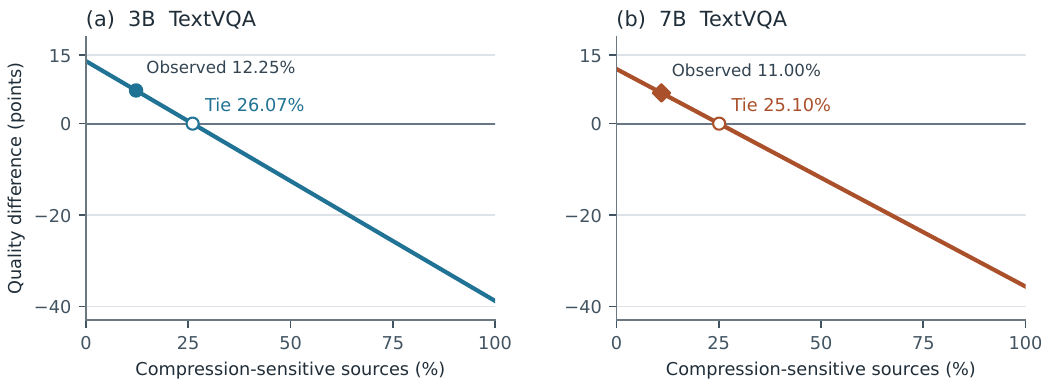}
\caption{\textbf{Which table wins as sensitive text images become more common?} The vertical axis is the TextVQA score of the population table minus that of the sensitive table. The filled marker is the confirmation share receiving full credit under Full and zero under D25. The open circle marks equal scores. Each line holds the two within-group means fixed.}\label{fig:mix}\end{figure}
Let $\rho$ be the sensitive share and $d_1,d_0$ the population table's advantages inside and outside the group. Holding within-group means fixed gives a TextVQA advantage of $\rho d_1+(1-\rho)d_0$. The two advantages are $-38.78$ and $+13.68$ points on 3B, and $-35.64$ and $+11.94$ on 7B. The sensitive group was defined by failure under D25, precisely the action chosen by $\su$. Its loss has a built-in direction, but the complementary group's advantage and the relative group sizes are empirical.

The tables tie at $\rho^*=d_0/(d_0-d_1)$, or 26.07\% ($[22.33,30.37]$) on 3B and 25.10\% ($[19.72,31.77]$) on 7B (Figure~\ref{fig:mix}). Only 245/2,000 and 110/1,000 confirmation images are sensitive, respectively. Development-only estimates place the crossings at 45.97\% and 23.12\%. A new service requires its own group means and prevalence to locate the crossing.

\subsection{Sensitivity to the evaluation subset}\label{sec:filter}
VTC-Bench selects images on which downsampling fails, while PAA-Bench uses repeated random pruning to characterize difficulty~\cite{vtc,paa}. In a retrospective check, we used a different model and operator to select images than to score the two competing methods. On 1,000 TextVQA images, define the gap as the 7B score with downsampling minus its score with DivPrune, both at the nominal one-quarter token budget. Across all images the gap was $-5.19$ points. Restricting to images on which 3B earned full credit with the original input gave $-3.90$. Among those images, the gap was $+12.64$ on the 91 for which 3B scored zero with one-quarter VisPruner and $-31.33$ on the 75 for which it scored zero with default CARES. The two sets shared 15 images. Changing the filter shifted the gap by $43.97$ points (pointwise 95\% bootstrap interval $[28.84,59.25]$, resampling image IDs jointly). The corresponding shifts on ChartQA and GQA were $+6.72$ and $-9.82$ points, with both intervals including zero. All three comparisons reuse the scored images. CARES chooses the resolution separately for each input, so filter identity and compression strength can both contribute.

\section{When does the task label improve the quality--time trade-off?}
\subsection{Complete requests at the deployment operating point}
We first decide from development data which tasks warrant a preview. Enabling it for neither task, only charts, only scene text or both gives development scores of 82.50, 81.13, 78.19 and 79.38 on 3B, and 84.825, 84.075, 84.800 and 82.000 on 7B. Under the development budget both selections return the population table, though the 7B margin over text-only preview is just 0.025 points.

We then time four workflows on the same 500 charts and 500 text images per model, with four scheduled repeats. The clock starts before opening the image and ends after answer decoding and device synchronization, including preview, dispatch, transfer and generation. Quality comes from these same timed answers (Table~\ref{tab:time}). Compared with $R$, $\su$ gains 3.50 points ($[1.52,5.48]$) and saves 12.57\,ms ($[2.86,22.28]$) on 3B. On 7B it gains 1.92 points ($[0.05,3.79]$) and saves 31.76\,ms ($[21.89,41.63]$). Default CARES has the highest mean quality among the four workflows and the longest mean time on both models.

\begin{table}[htbp]\centering\small
\caption{\textbf{Quality and complete-request latency on the same 1,000 images.} ChartQA and TextVQA have equal weight. Each model and workflow has four repeats. Bold marks the best observed value within a model and metric. The paired intervals in the text assess mean differences between the population table and the image router.}\label{tab:time}
\begin{tabular*}{\linewidth}{@{\extracolsep{\fill}}lrrr@{\hskip 1.25em}rrr@{}}\toprule
&\multicolumn{3}{c}{\textbf{Qwen2.5-VL-3B}}&\multicolumn{3}{c}{\textbf{Qwen2.5-VL-7B}}\\
\cmidrule(lr){2-4}\cmidrule(l){5-7}
\textbf{Workflow} & Score $\uparrow$ & Mean (ms) $\downarrow$ & P95 (ms) $\downarrow$ & Score $\uparrow$ & Mean (ms) $\downarrow$ & P95 (ms) $\downarrow$\\\midrule
Sensitive table &75.92&493.24&793.53&80.30&434.49&728.45\\
Population table &79.96&\textbf{478.66}&757.85&83.68&\textbf{414.74}&\textbf{665.30}\\
Image router &76.46&491.23&\textbf{732.09}&81.76&446.50&687.84\\
Default CARES &\textbf{80.31}&1311.65&1561.79&\textbf{85.66}&1202.56&1407.50\\\bottomrule
\end{tabular*}\end{table}

\begin{figure}[htbp]\centering
\includegraphics[width=.985\linewidth]{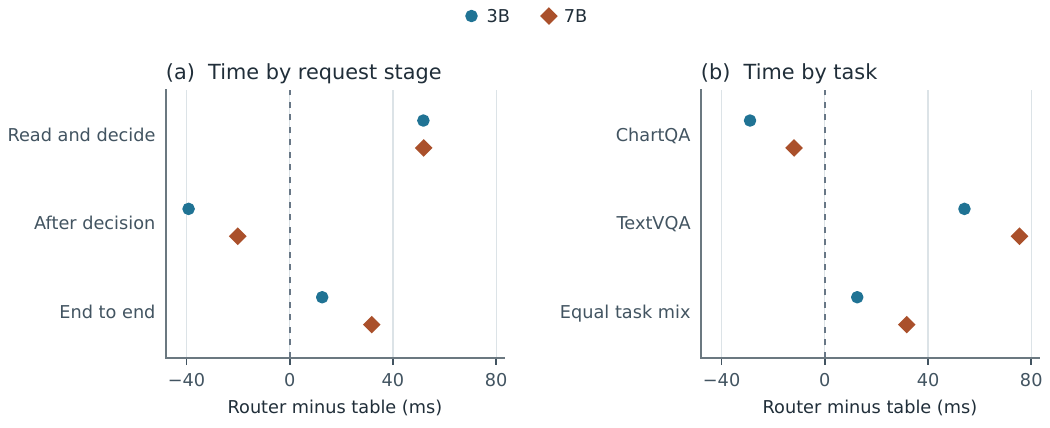}
\caption{\textbf{Where the preview gains or loses time.} Each value is the image router's mean time minus the population table's on the same images. Positive means the router is slower. (a) Read-and-decide time plus downstream execution equals the complete difference. (b) Charts favor the router, scene text favors the table. The equal-weight total is common to both panels.}\label{fig:time}\end{figure}
The stage times explain the mean. On 3B the router adds 51.78\,ms up to the decision and recovers 39.21\,ms downstream. On 7B it adds 51.90\,ms and recovers 20.14\,ms (Figure~\ref{fig:time}). At the frozen time prices the resulting complete-utility advantages for the table are 4.04 and 3.03 points (six-test Holm-adjusted $p=0.0292$ and $0.0402$). Matching falls short of the condition in Eq.~\ref{eq:breakeven} on this workload.

The mean combines opposing task effects. On 7B the router saves 11.91\,ms on charts and adds 75.44\,ms on scene text. Holding the within-task means fixed, it becomes faster above a chart share of 86.4\% (65.1\% on 3B). The 3B table also has a higher P95 despite its lower mean. Both population tables exceed their development budgets of 385.11 and 377.59\,ms on the complete-request clock.

\subsection{Equal target-action time under tighter budgets}
We next ask what the task label buys when both policies have the same mean time for target execution. The task-informed policy chooses among $7^2=49$ chart/text tables using four folds and applies the choice to the held-out fifth, rotating over all folds. Its comparator can use any \emph{shared} mixture of the seven actions, including Full. We give this comparator the evaluated action means and take its best quality at the table's realized cost on their upper convex frontier. Three measured sessions provide target-action times on 400 primary images for 3B and 267 for 7B. This clock excludes dispatch and preview.

\begin{figure}[htbp]\centering
\includegraphics[width=.985\linewidth]{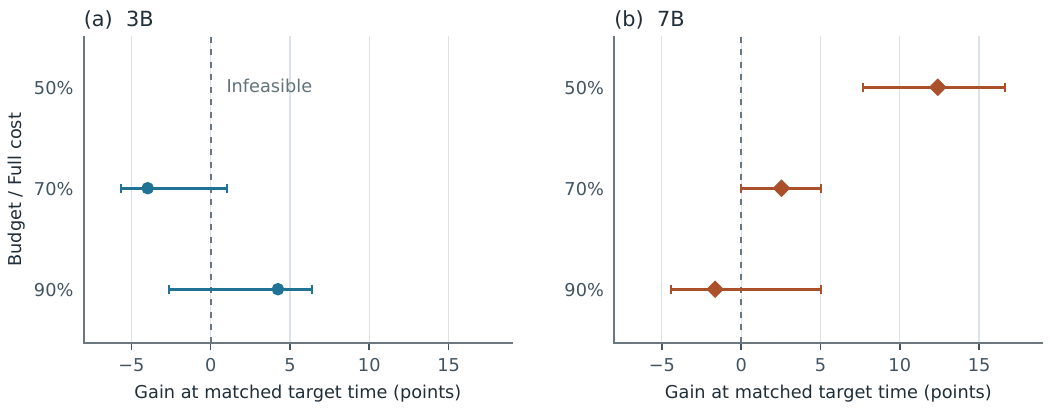}
\caption{\textbf{Quality gain over the best shared mixture at equal target-action time.} The nominal budget is a fraction of development Full cost. Positive values favor task-specific allocation. A cell is marked infeasible if a table cannot meet the budget in every fold. Pointwise 95\% intervals repeat five-fold selection and frontier fitting in 5,000 source resamples conditional on three timing sessions.}\label{fig:front}\end{figure}
At the 7B 50\% budget, upgrading D09 to D25 buys 42.86 points for 20.93\,ms on charts, versus 13.81 points for 44.70\,ms on text. All five folds therefore select D25 for charts and D09 for text. At a measured target-action time of 247.36\,ms this table scores 68.56, versus 56.16 for the best shared mixture, a paired gain of 12.39 points ($[7.69,16.64]$). The shared mixture uses D25 on 31.89\% of each task and D09 on the rest. Relative to it, the table moves 14.26\,ms from text to charts. Chart quality increases by 29.19 points and text quality decreases by only 4.40, explaining the equal-weight gain.

The same frontier can be read in the other direction. To match the table's quality the shared mixture requires 261.72\,ms, so the table saves 14.36\,ms ($[9.40,18.65]$). On 2,000 separate 7B confirmation sources the frozen table gains 12.66 points ($[11.51,13.80]$) at equal \emph{modeled} target-action time. This larger comparison measures answer quality directly and estimates time.

The 7B 50\% result emerged from a six-cell exploratory grid (Figure~\ref{fig:front}). At 50\% on 3B, no training fold admitted a feasible table. At 70\% and 90\% the intervals include zero on both models. Eq.~\ref{eq:task} describes a population optimum at a common price. A table learned from finite data can fall below the best shared mixture chosen with evaluation means. The simultaneous quality and mean latency gains for complete requests were measured separately in Table~\ref{tab:time}.

\section{Discussion}
The discrete router gets one decision right and another wrong on the same charts. Its selected images suffer less from downsampling than charts in general, which explains the positive matching term. Its forecasts on those very images still understate the loss, and paying for a preview leaves less budget for target execution. Repricing the preview changes the chosen allocation with the quality forecasts held fixed. Conversely, correcting the task-wide mean forecasts leaves the matching order for each action pair intact and fails to restore quality at the original budget. These controls suggest evaluating a redesign on the images it actually downgrades and measuring the frequencies it chooses after the preview is charged.

The appropriate comparator follows from the deployment decision. To improve a selector's image matching, exchange its existing recommendations and compare paired answers, as in the 7B CARES intervention. To decide whether to call it, run the alternative path on the same images. The observed text bypass, the mean versus P95 ranking on 3B, and the grid of target-action budgets each illustrate how a different operating condition can change the preferred policy. Time savings also depend on which operations a decision can skip~\cite{breakeven}. For a service with known task labels, a population-calibrated table provides a low-cost starting point. A preview is justified where its measured gain covers both the change in allocation and its net end-to-end cost.

Both sizes belong to the Qwen2.5-VL family. Complete requests were measured at batch size one on A100 PCIe 40\,GB GPUs. Task identification costs, batching and other devices can change the observed break-even point. The fixed-count oracle and sensitivity groups require answers that are unavailable online. Their role is to locate missed opportunities and workload dependence. The lookup table and the routers are the policies that can serve a new request.

We propose using the sensitivity of compressor rankings to the choice of evaluation images as a diagnostic for interpreting benchmark results. We will test how well this diagnostic guides the choice of a compression method for a target workload.

\section{Conclusion}
For the discrete visual router, the images assigned to compression were better chosen than chance, but too many charts were compressed for matching to recover the loss. A fixed-count reassignment makes both facts visible, and the same intervention detects useful resolution matching by CARES on 7B. When task labels are available, pairing this diagnostic with a timed task table shows whether image-specific decisions improve the service a user actually receives.

\input{references}
\appendix

\section{Sources, action settings and inference}
Table~\ref{tab:scale} distinguishes distinct sources from repeated answers and requests. In particular, the 32,000 timed observations are four workflows run in four scheduled repeats at each model size on the same 1,000 source images. All uncertainty estimates below retain the pairing of policies on a source.
\begin{table}[htbp]\centering\small
\caption{\textbf{Sources and measurements by experiment.} The 7B confirmation sources are drawn from the 3B pool. The action-timing row includes the auxiliary GQA workload. The final column counts generated answers or timed requests, including repeated measurements.}
\label{tab:scale}
\begin{tabular}{@{}lrrr@{}}\toprule
\textbf{Measurement} & \textbf{3B sources} & \textbf{7B sources} & \textbf{Answers or requests}\\\midrule
Development, seven actions & 12,000 & 3,000 & 105,000\\
Confirmation, seven actions plus CARES & 5,000 & 3,000 & 64,000\\
Action timing, seven actions, three sessions & 600 & 400 & 21,000\\
External resolution permutations, five draws & 600 & 600 & 6,000\\
Four complete workflows, four repeats & 1,000 & 1,000 & 32,000\\\bottomrule
\end{tabular}\end{table}

\subsection{How the fixed actions compare}
Table~\ref{tab:fixed} places the population table alongside the seven fixed actions. Full gives the highest quality on both sizes and D09 the shortest predicted complete time. The table gains quality and reduces predicted time relative to every fixed pruning setting in this implementation. The sharper task-information comparison in Figure~\ref{fig:front} therefore allows its shared baseline to mix all seven actions.
\begin{table}[htbp]\centering\small
\caption{\textbf{Quality and predicted complete time for each fixed action and the population table.} Scores use the primary confirmation answers. Bold marks the highest score or lowest estimate within a model. Table~\ref{tab:time} gives direct complete-request measurements.}
\label{tab:fixed}
\begin{tabular}{@{}lrrrr@{}}\toprule
&\multicolumn{2}{c}{\textbf{3B}}&\multicolumn{2}{c}{\textbf{7B}}\\
\cmidrule(lr){2-3}\cmidrule(l){4-5}
\textbf{Input policy} & Score $\uparrow$ & Est.\ time (ms) $\downarrow$ & Score $\uparrow$ & Est.\ time (ms) $\downarrow$ \\\midrule
Full &\textbf{86.79}&508.43&\textbf{91.90}&524.35\\
D25 &69.41&296.95&76.40&269.05\\
D09 &39.51&\textbf{266.76}&47.09&\textbf{234.68}\\
DivPrune $1/4$ &71.96&520.55&80.50&501.18\\
DivPrune $1/9$ &58.51&506.81&61.81&487.47\\
VisPruner $1/4$ &74.68&521.21&80.47&504.27\\
VisPruner $1/9$ &57.33&517.36&58.59&503.32\\
Population table &80.11&373.35&85.10&358.38\\\bottomrule
\end{tabular}\end{table}
\subsection{Inference and score checks}
OpenAI Codex assisted with literature searches, analysis code, and manuscript preparation.

Sources with matching pixel fingerprints or with a near-duplicate link to development images are excluded from confirmation. Target inference uses bfloat16, greedy decoding with one beam, a 64-token output cap and a unit repetition penalty. Full uses a processor budget of 4 to 1,280 visual tokens subject to dimension rounding. For the base response predictor, we use five paired initializations, AdamW at $3\times10^{-4}$, batch size 64, and choose the number of updates (50 to 1,000) on an internal training split. The primary deployment seed is zero. The low-capacity quality update uses a fixed 64-dimensional projection of image and text features. The default CARES path uses its released SmolVLM-500M-based selector. A secondary ChartQA scoring check that normalizes spaces within numerals changes the table-over-router difference from 5.70 to 5.20 points on 7B and leaves the 3B difference at 8.40 points.

\section{Uncertainty, timing protocol and additional controls}
\subsection{Paired inference and interrupted timing sessions}
We compare quality on paired sources within each task and aggregate with fixed task weights. The frozen population- versus sensitive-table contrast uses a paired $t$ interval with task-stratified variance. For the decomposition, 5,000 stratified source resamples recompute the frequencies and both quality terms. Complete-workflow mean differences combine source and scheduled-session variance using a four-session $t$ critical value.

Model loading and warmup precede the request clock. All accepted slow requests enter the mean. A violation of the preset background-resource rule triggers remeasurement of the complete image block. The first three 3B repeats each used one model load on one host. The fourth used four warmed segments on another host. The mean speed advantage of the population table was 12.33\,ms in the first three repeats pooled and 13.29\,ms in the fourth. All four 7B repeats used one load each on that second host. Direct cross-size latency differences also reflect the host and generated output.

\subsection{Prespecified tests and the calibration alternative}
The confirmatory contrasts are (i) population versus sensitive policy selection, (ii) a three-anchor calibration versus ordinary seven-action calibration with a $-0.5$-point noninferiority boundary, and (iii) $R$ versus the population table in quality minus development-priced complete time. The quality differences in (ii) are $-0.075$ ($[-0.486,0.336]$) and $+0.255$ ($[-0.525,1.035]$). The six-test Holm-adjusted $p$ is 0.0858 for each model under the prespecified testing convention. The anchor uses 1,536 action labels on 512 adaptation sources, compared with 3,584 for ordinary calibration on the same sources. Noninferiority is unresolved after multiplicity adjustment. For (iii), $R$ loses 4.04 and 3.03 utility points, six-test Holm $p=0.0292$ and $0.0402$. Directional claims for the exploratory budget grid use pointwise intervals.

\subsection{Can a task-independent mixture have a positive allocation gain?}
If the task baseline is optimal among task-dependent mixtures for the same population, clock and budget, every feasible marginal policy has $A\leq0$. A separate four-level experiment shows what happens when the nominal budget leaves the realized cost unequal. The simple mixture gained 3.027 allocation points over a development-selected table and used 29.784\,ms more cached action time. Re-evaluating the old choice on the new frontier accounted for $+0.213$ points. Moving along the new frontier to the mixture's cost accounted for $+2.897$, and the mixture fell $0.083$ short of that frontier. These terms sum to $+3.027$ up to rounding. Most of the apparent allocation advantage here came from using more action time.

Compared directly with this mixture, the input-cost router gained $0.516$ more from matching ($[0.057,0.997]$) but lost $2.330$ in allocation. On 750 images with three online repeats, interpolating between adjacent measured endpoints to equalize mean complete time gave mixture advantages of $+1.17$, $-0.47$ and $-0.51$ points, with all paired intervals including zero. This four-level case uses its own action menu and clock.

\subsection{Resolution expansion on GQA}
On 1,000 GQA confirmation images per model, default CARES increases token count above Full on 956 images, with means of 755.10 and 269.47 visual tokens for CARES and Full. CARES versus Full scores are 65.2 versus 69.0 on 3B and 61.8 versus 68.3 on 7B. An image resizer can increase computation when its preferred size exceeds that of the reference processor.
\end{document}

%% file: layout.tex
\usepackage[a4paper,left=27mm,right=27mm,top=23mm,bottom=25mm]{geometry}
\usepackage{amsmath,amssymb,booktabs,graphicx,xcolor}
\usepackage[font=small,labelfont=bf,labelsep=period,justification=raggedright,singlelinecheck=false]{caption}
\usepackage{hyperref}
\definecolor{inkblue}{HTML}{19516A}
\hypersetup{colorlinks=true,linkcolor=inkblue,citecolor=inkblue,urlcolor=inkblue,pdfauthor={Chenxi Wang and Bin Wang}}
\newcommand{\su}{s_{\mathrm{U}}}
\newcommand{\sa}{s_{\mathrm{A}}}
\newcommand{\papertitle}[1]{\begin{center}\vspace*{2mm}
{\LARGE\bfseries #1\par}\ifx\paperauthors\empty\else\vspace{9pt}{\large\paperauthors\par}\fi
\end{center}\vspace{3mm}}
\newcommand{\paperauthors}{Chenxi Wang\qquad Bin Wang\textsuperscript{*}\\[2pt]{\normalsize Xi'an Jiaotong University}\\{\small \textsuperscript{*}Correspondence: bin.wang@xjtu.edu.cn}}